# Advancing Person Re-Identification: Tensor-based Feature Fusion and Multilinear Subspace Learning

1st Akram Abderraouf Gharbi
*dept. Electrical Engineering*
*University Mohamed Khider of Biskra*
Biskra, Algeria
akram.gharbi@univ-biskra.dz

2nd Ammar Chouchane
*University Center of Barika*
Amdoukal Road
Barika,05001 Algeria
ammar.chouchane@cu-barika.dz

3rd Abdelmalik Ouamane
*dept. Electrical Engineering*
*University Mohamed Khider of Biskra*
Biskra, Algeria
ouamane.abdelmalik@univ-biskra.dz

4th Mohcene Bessaoudi
*dept. Electrical Engineering*
*University Mohamed Khider of Biskra*
Biskra, Algeria
bessaoudi.mohcene@gmail.com

*Abstract*—Person re-identification (PRe-ID) is a computer vision issue, that has been a fertile research area in the last few years. It aims to identify persons across different non-overlapping camera views. In this paper, We propose a novel PRe-ID system that combines tensor feature representation and multilinear subspace learning. Our method exploits the power of pre-trained Convolutional Neural Networks (CNNs) as a strong deep feature extractor, along with two complementary descriptors, Local Maximal Occurrence (LOMO) and Gaussian Of Gaussian (GOG). Then, Tensor-based Cross-View Quadratic Discriminant Analysis (TXQDA) is used to learn a discriminative subspace that enhances the separability between different individuals. Mahalanobis distance is used to match and similarity computation between query and gallery samples. Finally, we evaluate our approach by conducting experiments on three datasets VIPeR, GRID, and PRID450s.

*Index Terms*—Person Re-Identification, Multilinear subspace learning, Tensor feature fusion, pre-trained CNN, LOMO, GOG, TXQDA.

## I. INTRODUCTION

Today, the video surveillance system has wide pervasion and has big significance in security applications. Person re-identification is an essential task in video surveillance systems, and it is one of the most important topics in the computer vision field. The goal of person PRe-ID is to detect the similarity between pedestrian images sorted from non-overlapping cameras. Several problems exist with PRe-ID systems due to various factors, such as low resolution, occlusion, illumination variation, and pose changes [1], [3], [18], [38], [43], [44], [45]. The PRe- ID system contains three principal phases: feature extraction, learning, and matching. The proposed techniques in our issue are the same methods that were used in the other computer vision tasks, to achieve the best scores, many works propose different types of descriptors for robust data representation. Some of those descriptors focus on local features like color, texture, edges, and contours, for example LOMO [2], GOG [4], LDFV [7], HSCD [8], TPLBP [38], LBP, LBQ and BSIF [10], [46], [50]. Some others use deep learning and transfer learning models to extract the deep features [11]–[14], these approaches aim to enhance feature representation to improve the performances. Another side, some works propose methods to improve the metric learning stage like [2], [16], [17], [19]–[21], [26], [40], [41], [42]   by increasing the between class and decreasing the within class. For instance, Lin et al. [22] proposed Modified Linear Discriminant Analysis (LDA) [49] where the ficher vectors integrated with deep neural networks for learning nonlinear representations of person images in a space where the data is linearly separable. Jieru et al. [23] proposed Geometric Pre- serving Local Fisher Discriminant Analysis (GeoPLFDA). The method integrates the LFDA discriminative framework with a geometric conservation method that uses nearest-neighbor graphs to approximate local manifolds. Liao et al [2] proposed a new subspace metric learning technique named Cross-view Quadratic Discriminate Analysis (XQDA). Indeed, XQDA is an extension of the Bayesian face and KISSME [24] methods to cross-view features metric learning.

Multilinear Subspace Learning using tensor data has emerged as a powerful approach for person re-identification tasks. Traditional methods often treat visual data as a matrix, disregarding the inherent multi-modal nature of the data [15], [25], [47], [51]. However, by leveraging tensor representations, we can effectively capture the complex correlations between different visual modalities such as color, texture, and shape, resulting in improved person re-identification accuracy [6], [26].

In Multilinear Subspace Learning, tensors are utilized to model the multi-modal features extracted from person images or videos. These tensors capture the joint variations across multiple dimensions, allowing for a more comprehensive understanding of the underlying data structure. By leveraging tensor decomposition techniques, such as Canonical Polyadic (CP) decomposition [27] or Tucker decomposition [28], we can extract low-dimensional subspaces that effectively preserve the discriminative information across different modalities.

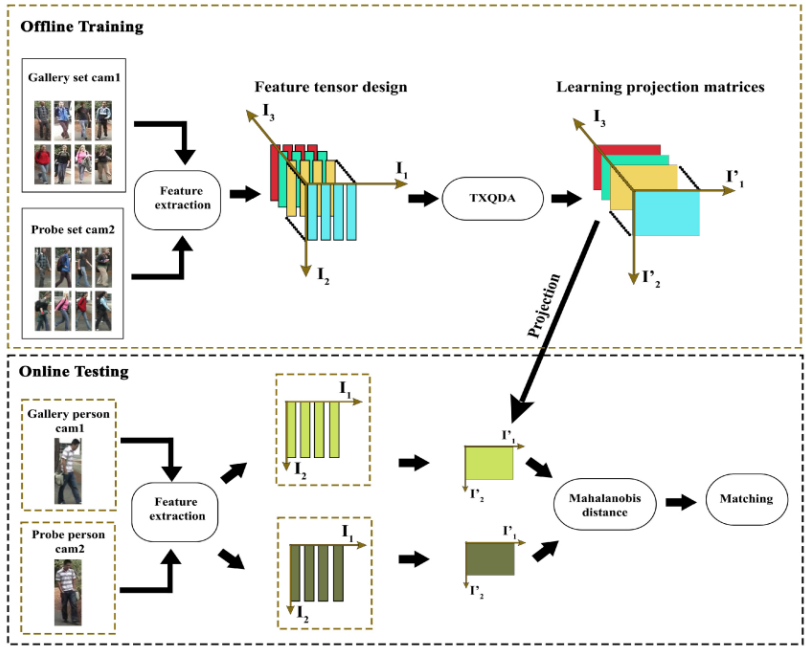

Fig. 1. The proposed approach pipeline of Person Re-id

Furthermore, Multilinear Subspace Learning enables the exploitation of the multilinear relationships among different modalities. By incorporating tensor-based fusion techniques, such as Tucker fusion or Kruskal fusion [29], we can effectively integrate the information from multiple modalities to enhance the discriminative power of the re-identification system. This fusion of tensor-based representations facilitates a more comprehensive and holistic representation of person images or videos, leading to more robust and accurate person re-identification [30].

Our contributions in this paper are summarized in the following:

- We propose a new multilinear representation of features based on pretrained Convolutional Neural Networks (CNNs) as a high-level feature extractor, merged with two local features descriptors Local Maximal Occurrence (LOMO) and Gaussian Of Gaussian (GOG). The fusion of these features in a tensor representation allows for capturing both spatial and semantic information, leading to improved discriminative capabilities.
- Additionally, we propose a new multilinear subspace projection algorithm named (Tensor-based Cross-View Quadratic Discriminant Analysis). TXQDA effectively exploits the inter-camera correlations and inter-modal interactions present in the tensor feature representation. By modeling the data in a multilinear framework, it learns a discriminative subspace that enhances the separability between different individuals.
- To evaluate the proposed method, we conducted experiments on three datasets, VIPeR, GRID, and PRID450s. The matching-based Mahalanobis distance is employed for similarity computation between query and gallery samples. The experimental results demonstrate the effectiveness of our approach, achieving significantly improved re-identification performance compared to existing methods. The obtained results affirm the potential of tensor-based fusion and multilinear subspace learning techniques in advancing the field of person re-identification.

The rest of this paper is organized as follow, the proposed methodology is provided in Section II, in which we describe the Tensor feature representation scheme and the Multilinear subspace learning process. Then, the Results and discussions are presented in Section III. Finlay, conclusion and future directions are given in Section IV.

## II. METHODOLOGY

### A. Proposed Person Re-Identification approach

The proposed approach of Person Re-identification (PRe-ID) system is shown in Fig1. Generally, the mechanism of Person Re-identification is to detect similar images of the candidate person through the gallery image datasets of various cameras. So, the system ranks the images of persons relying on the similarity with the probe. Based on the previous, our system includes of three essential stages: Design the Tensor of the extracted features, multilinear subspace learning, and matching with score normalization.

TABLE I
OUR SYSTEM PERFORMANCE RESULTS ON VIPeR

| Features types | Dim | Matching rates (%) | | | |
|---|---|---|---|---|---|
| | | Rank-1 | Rank-5 | Rank-10 | Rank-20 |
| CNN+LOMO | 50 | 43.73 | 76.27 | 87.37 | 94.84 |
| | 100 | 49.81 | 80.95 | 90.44 | 95.92 |
| | 150 | 52.12 | 82.69 | 90.82 | 96.20 |
| | 200 | 52.31 | 83.64 | 90.7 | 96.11 |
| | 250 | 53.16 | 83.32 | 90.19 | 95.82 |
| CNN+GOG | 50 | 39.78 | 74.59 | 85.95 | 93.01 |
| | 100 | 46.42 | 79.15 | 89.24 | 94.56 |
| | 150 | 48.64 | 81.46 | 89.56 | 94.84 |
| | 200 | 48.92 | 81.30 | 89.27 | 94.53 |
| | 250 | 49.72 | 80.82 | 89.30 | 94.53 |

TABLE II
OUR SYSTEM PERFORMANCE RESULTS ON GRID

| Features types | Dim | Matching rates (%) | | | |
|---|---|---|---|---|---|
| | | Rank-1 | Rank-5 | Rank-10 | Rank-20 |
| CNN+LOMO | 50 | 86.48 | 87.04 | 87.68 | 88.72 |
| | 100 | 86.48 | 87.20 | 88.00 | 88.48 |
| | 150 | 86.32 | 87.12 | 87.84 | 88.72 |
| | 200 | 86.32 | 87.20 | 88.00 | 88.80 |
| | 250 | 86.24 | 87.12 | 87.92 | 88.88 |
| CNN+GOG | 50 | 86.56 | 87.12 | 87.92 | 89.12 |
| | 100 | 86.40 | 87.20 | 87.68 | 89.28 |
| | 150 | 86.32 | 87.20 | 88.00 | 89.12 |
| | 200 | 86.48 | 87.20 | 88.08 | 89.68 |
| | 250 | 86.48 | 87.36 | 88.16 | 89.44 |

TABLE III
OUR SYSTEM PERFORMANCE RESULTS ON PRID450S

| Features types | Dim | Matching rates (%) | | | |
|---|---|---|---|---|---|
| | | Rank-1 | Rank-5 | Rank-10 | Rank-20 |
| CNN+LOMO | 50 | 62.53 | 90.09 | 95.38 | 98.27 |
| | 100 | 67.73 | 92.76 | 96.53 | 98.80 |
| | 150 | 69.82 | 93.38 | 96.40 | 98.80 |
| | 200 | 70.40 | 93.64 | 96.44 | 98.76 |
| | 250 | 69.78 | 93.02 | 96.22 | 98.67 |
| CNN+GOG | 50 | 57.73 | 87.07 | 94.00 | 98.00 |
| | 100 | 64.80 | 90.49 | 96.04 | 98.76 |
| | 150 | 66.18 | 91.42 | 96.36 | 98.84 |
| | 200 | 67.20 | 91.91 | 96.67 | 98.93 |
| | 250 | 68.09 | 91.82 | 96.71 | 99.02 |

### B. Tensor feature representation

To extract the image features, Three descriptors are used CNN [31] for deep features, LOMO [2], and GOG [4] for shallow features to produce three features vectors for each person image of the gallery, these descriptors are very effective in low- resolution, lighting, viewpoint, and background variations. For robust representation, each vector splits into parts to create a 3-order tensor, each tensor has three modes, the first one represents the number of feature parts, the second mode is the features, and the third mode represents the persons. Then, we combine the tensors to get CNN+LOMO Tensor and CNN+GOG Tensor. Fig2 illustrates the technique of feature extraction and tensor design.

In the offline training phase, the proposed technique TXQDA projects the training tensors X and Y on a new discriminant subspace, and the dimensions of both tensors are reduced to obtain new dimensions $I_1^{'} \times I_2^{'}$ for mode-1 and mode-2 respectively, where $I_1^{'} \times I_2^{'} \ll I_1 \times I_2$, mode-3 remains the same because represents the persons in the database. The following algorithm 1 depicts the differents step of TXQDA algorithm. In the online testing phase, the same procedures occur on each probe of pair images. After the projection, Mahalanobis distance [32] is computed to conduct the matching of the probe and the gallery in the new discriminant subspace. subspace.

### III. RESULTS AND DISCUSSION

The experimental results of the used databases VIPeR, GRID, and PRID450s are shown in Tables I, II, and III respectively, and their CMC curves are illustrated in Fig.3, Fig.4 and Fig.5. We use the 10-folder cross-validation protocol [33] in the experiments and evaluate the performance of our

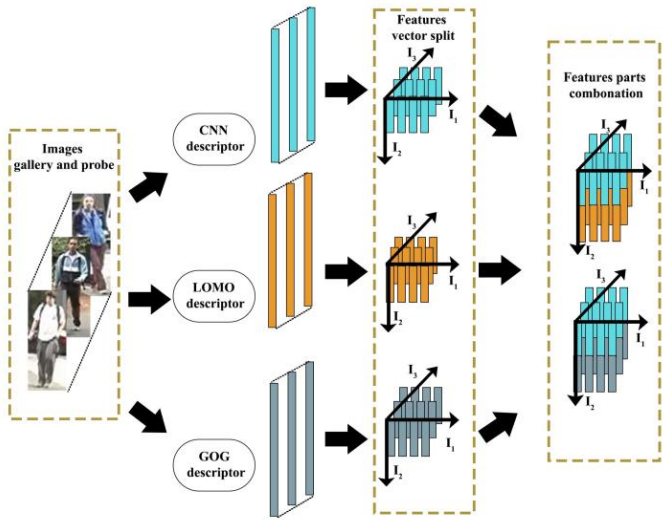

Fig. 2. The tensors design process

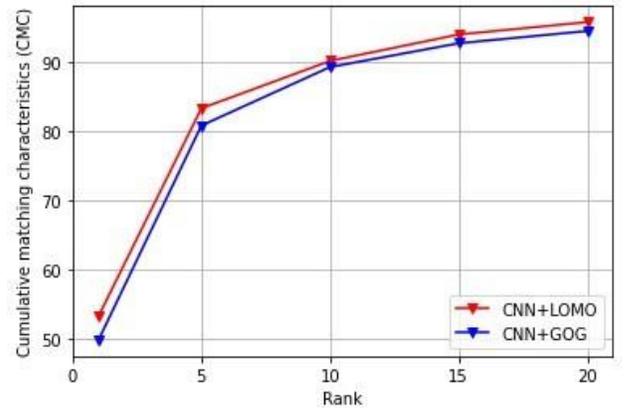

Fig. 3. CMC curves of the best scores on VIPeR database.

**Algorithm 1:** TXQDA

1 **Input:**
- $X$ and $Y$, the cross-view (Camera1-Camera2) training tensors.
- max_itr the maximum number of iterations.
- The projection dimensions $n'_1, n'_2, ..., n'_m$

**Initialization:** $P_1^1 = I_{n_1}, P_2^1 = I_{n_2}, ..., P_m^1 = I_{n_m}$ where I is the identity matrix.

**for** *iteration= 1 to* max_itr **do**
    **for** *k=1 to m* **do**
- $X' = X \times_1 P_1^{itr-1} ... \times_{k-1} P_{k-1}^{itr-1} \times_{k+1} P_{k+1}^{itr-1} ... \times_m P_m^{itr-1}$
- Deploy the tensor $X'$ to obtain its unfolding matrices $X^k$ (k=1,2,... m)
- $Y' = Y \times_1 P_1^{itr-1} ... \times_{k-1} P_{k-1}^{itr-1} \times_{k+1} P_{k+1}^{itr-1} ... \times_m P_m^{itr-1}$
- Deploy the tensor $Y'$ to obtain its unfolding matrices $Y^k$ (k=1,2,... m)
- Step1: Calculate the covariance matrices $Q_w$ and $Q_b$
- Step2: Calculate the projection matrices $P^k$ in k-mode by eigenvectors decomposition.
- Calculate the transformation matrix P
- **If** iteration $> 2$ and $\left\| P_K^{TXQDA^{itr}} - P_K^{TXQDA^{itr-1}} \right\| < n_k \times n_k \times \epsilon, k = 1, ..., m$, **break**

    **End For**
**End For**

**Output:**
The k-mode projection matrices $P_K^{TXQDA^{max_{itr}}}, k = 1, 2, ..., m$.

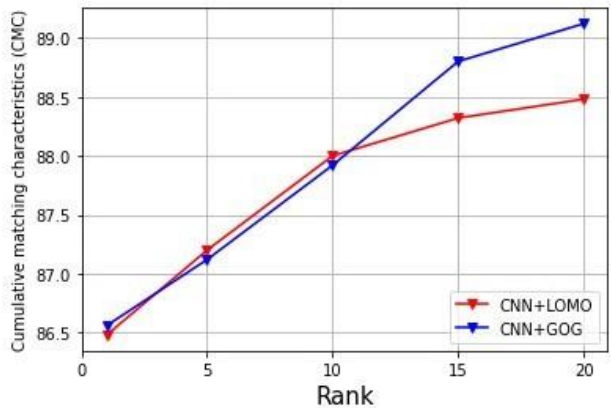

Fig. 4. CMC curves of the best scores on GRID database

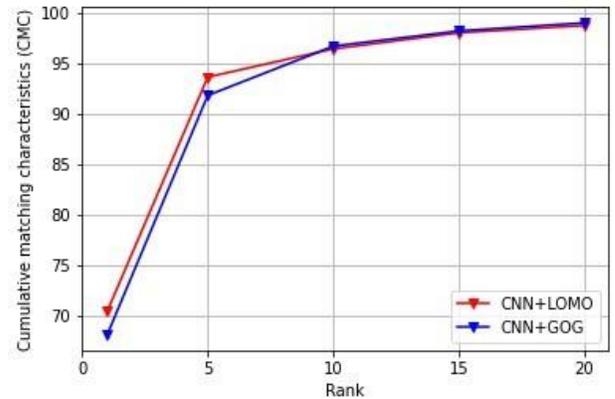

Fig. 5. CMC curves of the best scores on PRID450s database

approach using the cumulative matching characteristics (CMC) curves. The results into rank-1, rank-5, rank-10, rank-15, and rank-20 are given.

### A. Tensors features comparison

Generally, CNN+LOMO tensor has achieved the best results against CNN+GOG tensor in two datasets VIPeR and PRID450S but in GRID dataset CNN+GOG exceed CNN+LOMO by a few bit, and we note that TXQDA output dimension has a role in the variation of the scores percentages. In PRID450s dataset, the rank-1 of CNN+LOMO tensor reached 70.40% in dimension 200 and decreased in dimension 250, while the rank-1 of CNN+GOG tensor reached its best result 68.09% in dimension 250. In VIPeR dataset, the two tensors reached the best result in dimension 250 where CNN+LOMO achieved 53.16% and CNN+GOG achieved 49.72%. The only dataset in which CNN+GOG has a result better than CNN+LOMO is GRID, which has achieved 86.56% in dimension 50. In GRID dataset, we have achieved the best results in comparing the other previous works, CNN+LOMO obtained 86.48% in dimension 100 while CNN+GOG arrived at 86.56% in dimension 50.

### B. Comparison against the state-of-the-art

We have compared our method with previous work methods in recent years, and have achieved the best scores results in the used datasets. Table IV shows our technique against Person Re-ID state-of-the-art from rank-1 to rank-20.

## IV. CONCLUSION

In this paper, we propose a novel approach for person re-identification by combining tensor feature representation and multilinear subspace learning techniques. Our method leverages the power of pretrained CNNN as a high-level feature extractor, along with two complementary descriptors, LOMO and GOG. Furthermore, we incorporate a new multilinear subspace projection named TXQDA which is used to reduce the high dimension of the high-order tensor and create a new features representation with augmentation of the inter-class and detraction of the intra-class. As a future work, we propose to develop a new deep model trained by all the available datasets of Person re-identification in order to improve our system.

TABLE IV
PERFORMANCE COMPARISON WITH THE SOTA OF RANK-1 AND RANK-20 (%) ON VIPER, GRID, AND PRID450S DATASETS.

| Approach | Year | VIPeR | | GRID | | PRID450s | |
|---|---|---|---|---|---|---|---|
| | | Rank-1 | Rank-20 | Rank-1 | Rank-1 | Rank-20 | Rank-20 |
| FT-CNN+XQDA [11] | 2016 | 42.50 | 92.00 | 25.20 | 64.60 | 58.20 | 94.30 |
| SSDAL+XQDA [34] | 2016 | 43.50 | 89.00 | 22.40 | 58.40 | 22.60 | 69.20 |
| Kernel X-CRC [35] | 2019 | 51.60 | 95.30 | 26.60 | 69.7 | 68.80 | 98.40 |
| VS-SSL [36] | 2020 | 43.90 | 87.80 | - | - | 63.30 | 97.00 |
| SLDL [37] | 2022 | 51.23 | 95.02 | 33.46 | 75.44 | - | - |
| TXQDA(CNN+LOMO)(Our) | 2023 | 53.16 | 95.82 | 86.48 | 88.72 | 70.40 | 98.76 |
| TXQDA(CNN+GOG)(Our) | 2023 | 49.72 | 94.53 | 86.56 | 89.12 | 68.09 | 99.02 |